\documentclass[letterpaper, 10 pt, conference]{ieeeconf}  % Comment this line out if you need a4paper

\IEEEoverridecommandlockouts                              % This command is only needed if 
\usepackage{graphics} % for pdf, bitmapped graphics files
\usepackage{epsfig} % for postscript graphics files
\usepackage{mathptmx} % assumes new font selection scheme installed
\usepackage{times} % assumes new font selection scheme installed
\usepackage{amsmath} % assumes amsmath package installed
\usepackage{amssymb}  % assumes amsmath package installed
\usepackage{helvet} % DO NOT CHANGE THIS
\usepackage{courier}  % DO NOT CHANGE THIS
\usepackage[hyphens]{url}  % DO NOT CHANGE THIS
\usepackage{graphicx}  % DO NOT CHANGE THIS
\usepackage{algorithm}
\usepackage[noend]{algpseudocode}

\title{\LARGE \bf
3DMotion-Net: Learning Continuous Flow Function for 3D Motion Prediction
}
\author{Shuaihang Yuan$^{*}$,
        Xiang Li$^{*}$,
        Anthony Tzes,
        Yi Fang$^{\dagger}$
        \\
NYU Multimedia and Visual Computing Lab, USA \\
New York University Abu Dhabi, UAE\\
New York University, USA \\
\thanks{* Those authors contributed equally to this paper. $\dagger$ Yi Fang is the corresponding author.}
}

\begin{document}
\maketitle
\thispagestyle{empty}
\pagestyle{empty}
%%%%%%%%%%%%%%%%%%%%%%%%%%%%%%%%%%%%%%%%%%%%%%%%%%%%%%%%%%%%%%%%%%%%%%%%%%%%%%%%
\begin{abstract}
In this paper, we deal with the problem to predict the future 3D motions of 3D object scans from previous two consecutive frames. Previous methods mostly focus on sparse motion prediction in the form of skeletons. While in this paper we focus on predicting dense 3D motions in the from of 3D point clouds. To approach this problem, we propose a self-supervised approach that leverages the power of the deep neural network to learn a continuous flow function of 3D point clouds that can predict temporally consistent future motions and naturally bring out the correspondences among consecutive point clouds at the same time. More specifically, in our approach, to eliminate the unsolved and challenging process of defining a discrete point convolution on 3D point cloud sequences to encode spatial and temporal information, we introduce a learnable latent code to represent the temporal-aware shape descriptor which is optimized during model training. Moreover, a temporally consistent motion Morpher is proposed to learn a continuous flow field which deforms a 3D scan from the current frame to the next frame. We perform extensive experiments on D-FAUST, SCAPE and TOSCA benchmark data sets and the results demonstrate that our approach is capable of handling temporally inconsistent input and produces consistent future 3D motion while requiring no ground truth supervision.
\end{abstract}

%%%%%%%%%%%%%%%%%%%%%%%%%%%%%%%%%%%%%%%%%%%%%%%%%%%%%%%%%%%%%%%%%%%%%%%%%%%%%%%%
\section{INTRODUCTION}
3D motion prediction has been a long-standing problem in computer vision and robotics, serving as an essential building block of many applications, such as object tracking, 3D rigging, virtual and augmented reality, human-robot interaction, and motion synthesis \cite{van2007augmented,zhou2008trends,gururajan2006gaming,javed2002tracking,coifman1998real,azarbayejani1996real,strauss1992object}. 
However, this task is still challenging due to sensor noise, occlusion, and dynamic topology changes, especially for real-world scans which exhibit significant shape deformations and variations.

%It is of great importance to develop a method that can predict and track consistent future motions from previous observations.
% With the rapid development of 3D scanners, it becomes easier for people to capture objects in many 3D representations, such as point clouds, voxel grids, and meshes in a Euclidean space. 

% In order to create a machine that can interact seamlessly with the world, it needs a similar ability to understand
% the dynamics of the human world, and to predict probable
% futures based on learned history and the immediate present.

% However, it is not practical to obtain a sequence of temporally consistent 3D activities scans due to background noises, occlusion, and dynamic topology changes. These challenging issues restrict the usage of those scans in real-world applications, especially in the application of assistive robots which is designed to perform precise real-time interaction with human, since it is of great challenge for robots to infer human future motions from data with noise, occlusion, etc. 

\begin{figure}[ht!]
\centering
  \includegraphics[width=\linewidth]{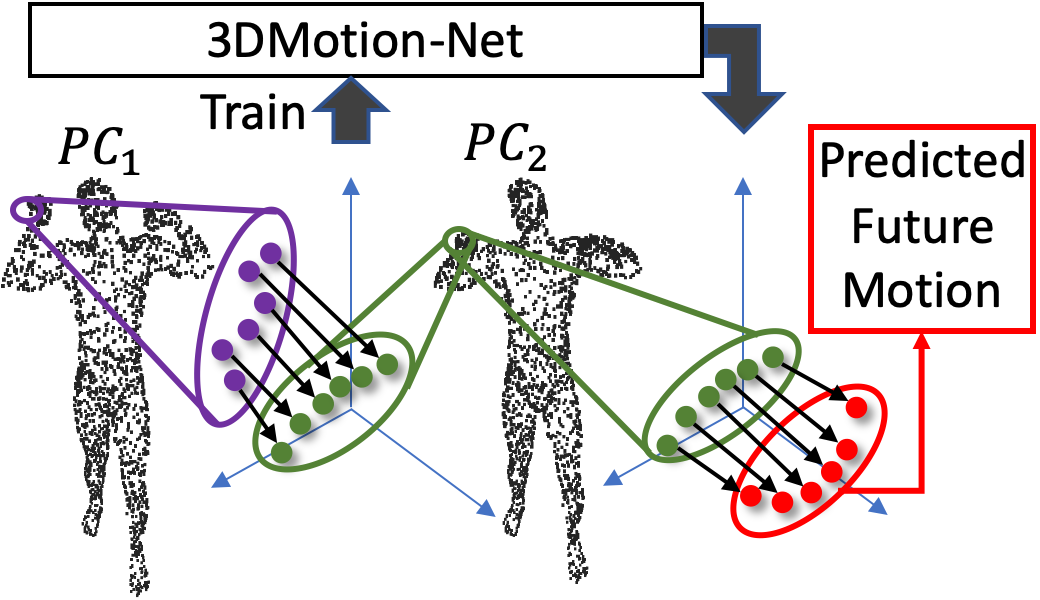}
  \caption{Illustration of 3D Motion prediction. Our model takes two consecutive point sets in 3D space as input and predicts the flow between two consecutive point sets. The learned motion patterns are used to infer the future motion by predicting the flow field from current frame to future one.}
  \label{fig:intro} 
\end{figure}
Existing motion prediction works mostly focus on predicting future object poses characterized by skeletons from a given past. Early efforts build their model with expert knowledge about motion patterns in the form of Markovian assumptions \cite{lehrmann2013non,pavlovic2001learning}, smoothness, or low dimensional embeddings \cite{wang2006gaussian}. Recent works use deep recurrent neural networks (RNNs) to better capture temporal characteristics \cite{martinez2017human}. Another family of methods leverages the power of generative adversarial networks (GAN) to enforce the predicted skeletons to be similar to a real sequence of poses \cite{barsoum2018hp}.

In contrast to previous researches, we aim at predicting dense 3D motions in the form of 3D point clouds rather than skeletons. Unlike skeletons that have pre-defined structures and with fewer node points, 3D point clouds are unstructured and unordered, which makes this task even more challenging. There are two main challenges for the task of dense 3D motions prediction from point clouds: 1) first, it's still a difficult task to learn robust and representative feature embeddings from 3D point clouds; 2) second, it's still unknown how to model the temporal correlation among consecutive frames. For the first challenge, researches have proposed extensive hand-crafted signatures \cite{rudenko2019human,huang2016volumetric,677048,kucner2017enabling,liao2003voronoi} to extract point features from 3D point clouds. Recent researches leverage the power of deep neural networks to encode the geometric information of input shapes to deep feature descriptors \cite{fang20153d}. %In these methods, they need to explicitly design a feature encoding network to extract high-level feature embeddings followed by a decoder network to transform these features to object tracking or correspondence results \cite{hesse2019learning}. 
Following the pioneering work of PointNet \cite{qi2017pointnet} that directly applies convolution on raw point clouds, various point convolutions operations have been developed to learn shape descriptors from 3D scans \cite{qi2017pointnet++,verma2018feastnet,le2018pointgrid}. Nevertheless, it's still challenging and non-trivial to find the best convolution operation that is powerful and robust enough to deal with real-world scans.

Regarding the second challenge of learning spatiotemporal correlations among consecutive frames, this is an area with less research attention. Previous learning-based methods mostly focus on modeling 3D motions of existing observations and can not deal with the problem of predicting future motions from previous observations. In addition to the spatial geometric features, the temporal features also play an important role in predicting the motions for time-varying 3D scans. However, it would be a challenging problem to define a convolution operation that is powerful enough to extract robust spatial and temporal features from time-varying scans. In \cite{niemeyer2019occupancy}, the authors propose probably the first spatiotemporal representation of time-varying 3D point clouds and verify the effectiveness for dense 4D reconstruction. Nevertheless, they still use PointNet-like encoder network for point feature learning of a sequence of point clouds.

\begin{figure*}[ht!]
\centering
  \includegraphics[width=\textwidth]{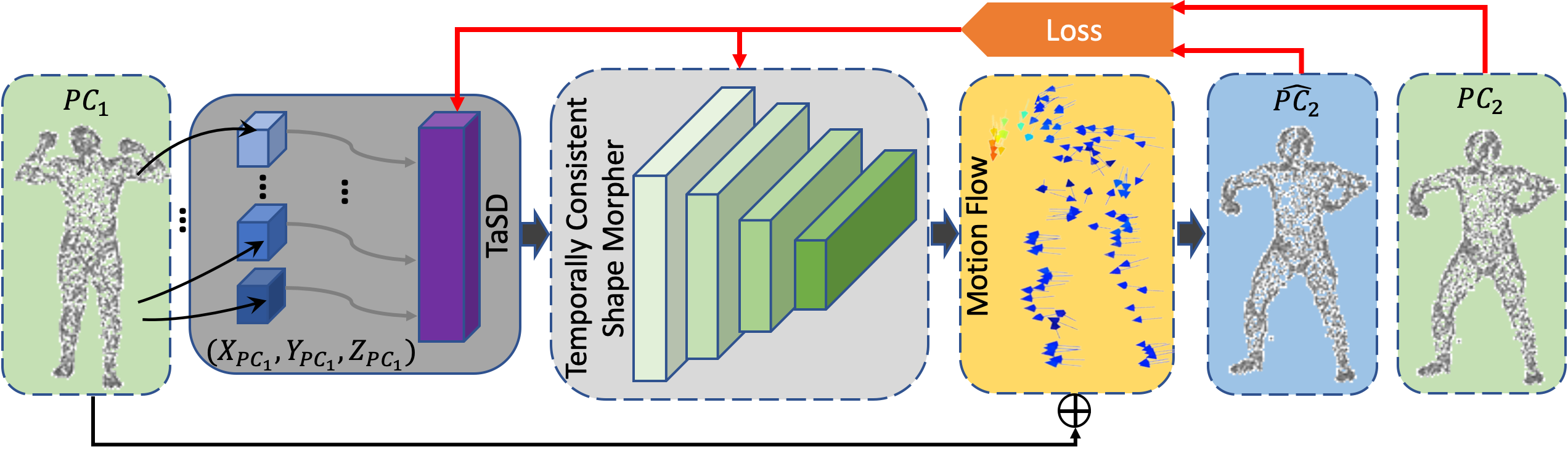}
  \caption{
The pipeline of the proposed method. We proposed two models to learn a continuous flow function that can predict the future motion. The first one is a temporally-aware shape descriptor that encoders the essence of spatiotemporal motion patterns between consecutive frames, shown in the purple box. The second module is temporally consistent shape Morpher that generates a 3D flow field from the current frame to the next frame, represented in the red block. Our module takes two consecutive point cloud $PC_1$ and $PC_2$ as inputs to train the Morpher and optimize the TaSD. We concatenate the TaSD to each point of $PC_1$ and feed it to the shape Morpher to obtain the predicted 3D motion. After applying the motion flow to $PC_1$ we get the predicted point cloud $\hat{PC_2}$, which is compared with $PC_2$ to formulate the loss term. During training, we optimize both shape Morpher network and the latent code $z$. While in the inference state, we fix the shape Morpher network and only optimize latent code $z$ towards the minimization of the loss between $\hat{PC_2}$ and $PC_2$ and then predict the future frame $\hat{PC_3}$.
  }
  \label{fig:pipline} 
\end{figure*}

\begin{figure}[ht!]
\centering
  \includegraphics[width=0.8\linewidth]{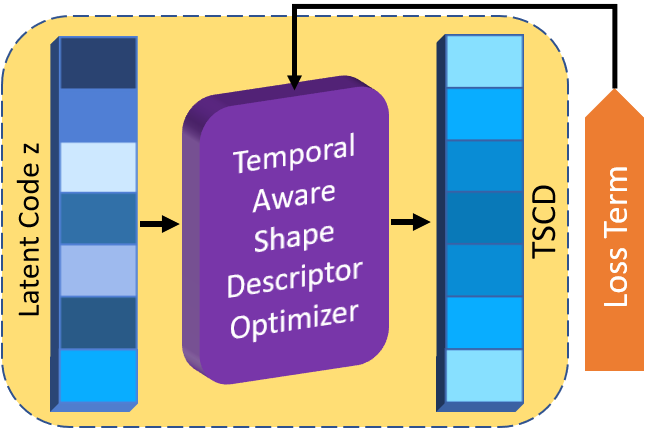}
  \caption{Illustration of TaSD Optimizer which learn the descriptor by optimizing a latent code $z$ using pre-defined loss. }
  \label{fig:optimizer} 
\end{figure}

In this paper, to avoid the challenging issue of spatiotemporal feature extraction using an encoder network, we introduce a learnable latent code to represent the temporal-aware shape features which can be optimized during model training. This temporal-aware shape representation is in nature captures both spatial geometry characteristics of 3D shapes and their correlation among consecutive frames. Our method is self-supervised which assuage the supervision of ground-truth correspondences and can predict the future motion and bring out correspondences of future motion naturally. The pipeline of our proposed method is presented in Fig.\ref{fig:pipline}. Three major models are enclosed to learn the desired 3d motion flow function. We first propose a temporal-aware shape descriptor optimizer, shown in Fig.\ref{fig:optimizer}that can learn the desired spatiotemporal representations implicitly. A temporally consistent motion Morpher, which takes as input a sequence of two 3D source shapes, is then used to deform the source shapes to the corresponding target shapes based on temporal-aware shape descriptor. This morphing module is built upon a deep neural network. Both temporal-aware shape descriptor and the morphing decoder network is optimized during the network training process using an unsupervised loss function. While in the test stage, the parameters of the decoder network are fixed, which can be regarded as learned prior information from the training data set, and we only optimize the temporal-aware shape descriptor towards the optimal spatiotemporal representation. The future motion with dense correspondence can be generated using the optimal spatiotemporal representation.

In summary, our main contributions are:
\begin{itemize}
\item We present a self-supervised method that can learn continuous flow function from a sequence of 3D scans. This function can predict the future object motions and naturally bring out the shape correspondences simultaneously. 
\item We introduce a novel spatiotemporal feature representation for time-varying 3D scans which eliminates the unsolved and challenging process of defining a discrete point convolution on continuous space to encode spatial and temporal information.
\item We conduct extensive experiments on several public data sets and demonstrate that our proposed method can be applied to temporally inconsistent scans to predict a sequence of temporally consistent motions.
\end{itemize}

\section{Related Works}
In this section, we first go through some feature learning techniques for 3D representations. Then, we summarize existing flow estimation methods and temporal feature learning approach.

\subsection{3D Feature Learning}
Though deep learning \cite{LeCun1998} has achieved significant progress in various image recognition tasks, the research on applying deep learning on raw point cloud to obtain a robust and informative feature is still challenging. There are multiple ways to represent 3D data, such as raw point clouds, voxels, meshes, leading researchers to different ways towards 3D signature learning.
Voxel Grid is a popular and adaptive data structure in 3D, which can be easily applied to 3DConvNet \cite{VOXel-deep,VOx-3DConv} to obtain voxel-wise features. The exceptional work of Wu \cite{VOXel-deep} firstly brings the success of CNN from 2D images into 3D. With a high-resolution volumetric representation of 3D Shape, most of the position and geometric information can be well preserved, which makes it possible to actively learn shape descriptors from the voxel-based representations. However, a high resolution of a voxel grid usually leads to cubically growing of space and computational cost. Some recent jobs focus on applying deep learning on meshes \cite{specOnMesh,SpecOnMesh02}. These methods, nevertheless efficient in manifold meshes, are hard to obtain a generalized feature that robust to structure variations. Multiview-based methods, e.g., \cite{Multiview-CNN}, transform 3D shapes into multiple 2D images and aggregates features of each 2D image to specialize in the feature of a 3D shape. Later work on multiview \cite{multiview02,multiview03}, extends this idea to multiple 3D tasks and show very persuasive performance. Even though those multiple views are very applicable and robust in some specified tasks, they are sensitive to the resolution of 3D data and they can only get information from the object surface. Among all 3D data types, point cloud, which is capable to render meticulous 3D models, is the most informative and rawest one. The information around a specified point and the information of the general object are usually refined as local features \cite{intrinsic02,intrinsic03} and global feature \cite{extrisic01,extrin03}. Due to its irregularity, most feature extraction approaches with point cloud are hand-crafted for a specific task. In \cite{qi2017pointnet}, the authors first introduce the network that directly consumes raw point clouds. PointNet uses a max-pooling layer to achieve invariant to different permutations of the point cloud. Extensive following works \cite{qi2017pointnet++,SoNet,le2018pointgrid,wang2019non} developed various point convolution operators to directly learn point signature from point clouds. Notable method PointNet++ \cite{qi2017pointnet++} groups points in different scales to extract the features hierarchically to obtain multi-scale point descriptors. Although these methods have achieved state-of-the-art performance on various point cloud processing tasks, they usually need to carefully design the point convolution operators and network architectures to ensure the quality of point descriptors. In contrast, DeepSDF \cite{park2019deepsdf} proposes an auto-decoder architecture that abandons the point feature encoding network and only use a decoder network to achieve 3D shape completion. A learnable latent code is designed to characterize the global feature representation, which is randomly initialized and optimized during network training. 

\subsection{3D Flow Estimation}
It is of great importance to study the problem of flow estimation on 3D point clouds in the robotics field. Many flow estimation of 3D point clouds using classical techniques has been well studied such as \cite{tanzmeister2014grid,ushani2017learning} which apply the occupancy maps to scenes and use particle filters \cite{danescu2011modeling} or EM to estimate the motion flow of the object. Besides the classical techniques, Dewan et al. \cite{dewan2016rigid} proposed to use the hand-crafted descriptor with a regulation term to estimator the rigid motion flow of a scan. In addition to the hand-crafted descriptors, researchers leverage the power of deep neural networks to learn the descriptor. PointFlowNet builds on the VoxelNet \cite{zhou2018voxelnet} to predict scene flow and rigid object motion. FlowNet3D \cite{liu2019flownet3d} uses the PointNet++ \cite{qi2017pointnet++} as the backbone to embed the flow. Instead of using PointNet++, HPLFlowNet \cite{gu2019hplflownet} use Bilateral Convolutional Layer (BCL) \cite{jampani2016learning} for a better global information extraction. Though previous works achieve a promising result for flow estimation, these methods focus on the finding explicit design of convolutional architectures to learn a better shape descriptor and fail to explore the temporal information between a sequence of inputs. In this work, we introduce a novel spatiotemporal feature representation for time-varying 3D scans which eliminates the unsolved and challenging process of defining a discrete point convolution on continuous space to encode spatial and temporal information.

\section{Problem Statement}
We leverage the power of deep neural networks to learn a continuous motion flow function for 3D motion prediction from consecutive frames of point clouds. Our model takes as input two sets of points at consecutive frames, which are randomly sampled from a time-varying 3D objects. Each point in the point sets has 3 dimensions which corresponding to the $(X,Y,Z)$ coordinates in the Euclidean space. It is flexible to use additional features such as point colors as the extra inputs but in this paper, we focus on the 3 dimensions inputs only.

Based on the settings above, we sample a set of points $P$ from one frame. For the next consecutive frame, we sample another set of points $Q$. Note that due to background noises, occlusion and dynamic topology changes, $P$, and $Q$ are not necessarily have corresponding relationships. Now, let us suppose a point $(x_i,y_i,z_i)$ is the $i^{th}$ point out of $N$ sampled points in the first frame, its corresponding location at the second frame $Q$ becomes $(x'_i,y'_i,z'_i)$. For this $i^th$ point at the first frame, the motion is represented by a point flow $f_i$, such that $f_i=(x'_i,y'_i,z'_i)-(x_i,y_i,z_i)$. Consequently, For the whole point set $P$, there exist a motion flow $F$ such that $F$ contains point flow for every point at the first frame to the next frame $\mathbf{F}=\{f_i\}_{i=1}^{N}$. In this paper, we aim to predict the motion flow from the second frame $Q$ to the future frame $Y$ driven by the existing motion flow from $P$ to $Q$.

\section{3DMotion-Net Architecture}
In this section, we introduce our motion flow prediction network, 3DMotion-Net, which alleviate the explicit design of encoder to extract spatial and temporal feature, see \ref{fig:pipline} for illustration. Our proposed method consists of two key components. The first component is a temporal-aware shape descriptor optimizer which tries to learn spatiotemporal representations, while the second component is temporally consistent motion Morpher which learns a continuous motion flow function that morphs a 3D scan from current frame to the next frame. We first present these two modules and then introduce the detailed network architecture.

\subsection{Shape Descriptor Learning with Temporal-Aware Shape Descriptor Optimizer}

Many previous works leverage the neural network to design a task-specific architecture for shape descriptor extraction. To apply discrete convolution to continuous 3D space especially to irregular and unordered point cloud, numerous point-based convolution operations, such as PointNet \cite{qi2017pointnet}, are proposed for a better point feature extraction. However, it is unclear whether there exists a best explicit feature extractor or it is necessary to use a complex extractor for feature extraction. This motivates us to reduce the dependency on the explicit design of feature extractor. We develop a novel learnable descriptor in latent space that requires no explicit feature extraction networks. 

% Generally, we group 3D coordinates of the observed motions with the descriptor to form a pair to predict a morphed shape.
As we showed in Fig.\ref{fig:pipline}, for every input pair of point sets $(P,Q)$, we assume that there exists a latent code $z_{PQ}$ that carries the geometric essences of two consecutive shapes as well as the translational relationship. To find the desired latent code for each input pair, we first randomly initialized this latent code from a Gaussian normal distribution, i.e., $z\sim \mathcal{N}(0,1)$. Then, the temporal-aware shape descriptor (TaSD) optimizer is applied to the initialized latent code to search for the optimum representation. We adopt the back-propagation strategy along with an unsupervised loss to optimize latent code $z$ to obtain the optimal descriptor. The loss function and the optimization rule of the $z$ are presented in the sections \ref{sc_loss}.

\subsection{Motion Flow Predicting with Temporally Consistent Shape Morpher}
\begin{figure}[ht!]
\centering
  \includegraphics[width=0.8\linewidth]{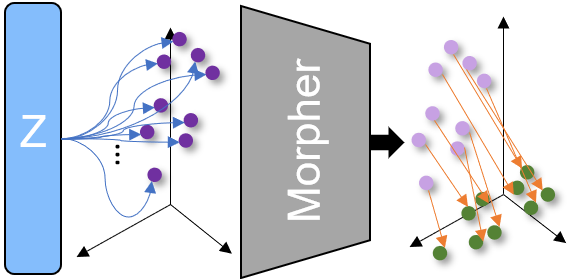}
  \caption{Illustration of temporally consistent shape Morpher to decode optimized TaSD to continuous flow field.}
  \label{fig:decoder} 
\end{figure}
To regress the flow from one frame to the next, we implement a deep neural network to learn the motion flow function $\mathcal{F}(\cdot)$. Given a pair of point sets $(P,Q)$ at two consecutive frames and the corresponding latent code $z_{PQ}$, our inputs are formulated as follows,
\begin{equation}
    X_{PQ}=(P,z_{PQ})
\end{equation}
. Then, we feed $X_{PQ}$ to the motion flow function to obtain motion flows as we presented in Fig.\ref{fig:decoder}.

We use a Multilayer Perceptrons (MLPs) with softplus activation function to approximate $\mathcal{F}(\cdot)$ using $\theta$ parameters. Hence our continuous motion flow function can be approximated as follows:
\begin{equation}
    \mathcal{F}(X_{PQ})\approx MLP_{\theta}([P,z_{PQ}])
    \label{eq1}
\end{equation}
, where $[\cdot,\cdot]$ represent the concatenation operation.
The motion flows can be obtained by the following equation:
\begin{equation}
    F_{PQ}=\mathcal{F}(X_{PQ})
\end{equation}, where $F_{PQ}$ is the predicted motion flows from $P$ to $Q$. Ideally, the predicted motion flow should be accurate enough such that point set $P+F_{PQ}$ is close to $Q$.

\subsection{Motion Flow Loss and Optimization}\label{sc_loss}
To ensure a good quality of predicted motion flow, we design a loss term $\mathcal{L}_{\theta,z}$ which is the summation of losses of two flows. For each flow, the loss is calculated by using the Chamfer distance:
\begin{multline}
    \mathcal{L}_{\theta,z_{PQ}}(P+F_{PQ},Q)=\frac{1}{|P+F_{PQ}|}\sum_{x\in P+F_{PQ}}\min_{y \in Q}\|x-y\|_{2}+ \\
    \frac{1}{|Q|}\sum_{x\in Q}\min_{y \in P+F_{PQ}}\|x-y\|_{2} 
    \label{eq:loss}
\end{multline}
. Note that Chamfer distance is calculated without using the ground truth correspondence information. This loss works as a key component in both training and inference state.

During training, our goal is to find the optimal network parameters $\theta$ and latent code $z_{PQ}$ such that:
\begin{equation}
     \mathop{\arg\min}_{\theta,z_{PQ}} \mathcal{L}_{\theta,z_{PQ}}(P+F_{PQ},Q)
    \label{eq:train}
\end{equation}
Note that both network parameters $\theta$ and latent code $z$ are updated simultaneously by this loss term during the training time.

In the inference state, we fix decoder parameters $\theta$ and reinitialize the latent code $z$ for each input pair. The learned network parameters $\theta$ can be regarded as prior information to guide the optimization process of latent code. During the test optimization process, since we do not know the future frame $Y$ during the inference, we use the loss term $\mathcal{L}_{\theta,z_{PQ}}(P+F_{PQ},Q)$ to optimize our latent code $z$ using the updating rule:
\begin{equation}
     \hat{z_{PQ}}=\mathop{\arg\min}_{z_{PQ}} (\mathcal{L}_{\theta,z_{PQ}}(P+F_{PQ},Q))
    \label{eq:opt}
\end{equation}
Once the loss converges, the prediction of a future motion can be inferred by applying the predicted motion flow to the second frame. Moreover, the correspondence can be obtained naturally after applying the motion flow to the original point set.

\section{Experiment}
\subsection{Data Sets, Implementation and Training Detail:}
We conduct experiments on several public data sets to show the abilities of our model in predicting 3D motions. Experimental data sets include Dynamic FAUST (D-FAUST), a dynamic human motion data set, which contains 40,000 meshes, a human pose data set SCAPE \cite{Scape} which consists of a sequence of real human scans, and the TOSCA \cite{tosca} data set which contains synthetic nonrigid models of 80 objects with various object category, such as cat, dog, wolf, etc. Note that although the ground truth corresponding points are provided for all these three data sets, our proposed method uses no ground truth information neither in the training nor optimization.  

The pipeline of our proposed method can be found in Fig.\ref{fig:pipline}. We generate training and testing episodes from the training and test set respectively, each with a length of 3 frames. For each frame, we randomly initialize a latent vector from a normal distribution, i.e.,$z\sim\mathcal{N}(0,1)$. By default, we set the dimension of $z$ to 256. The temporally consistent shape Morpher is implemented using an MLP with hidden layers of size 512, 256, 128, 64, and 3 to predict a flow field from the current frame to the next frame and the shape correspondences can be naturally generated by using this flow function. 

\subsection{Evaluation Metrics:}
We use the cumulative matching accuracy to evaluate the quality of the predicted future motion. For each point, it is considered as a correct match if the distance between this point and its corresponding ground-truth point is within a certain threshold $\Delta$. In addition to the cumulative matching accuracy, we also use chamfer distance and $\ell_2$-distance to evaluate the quality of prediction.

\subsection{Motion Prediction on Consistent data set}
\begin{figure}[!htp]
\centering
\includegraphics[width=0.85\linewidth]{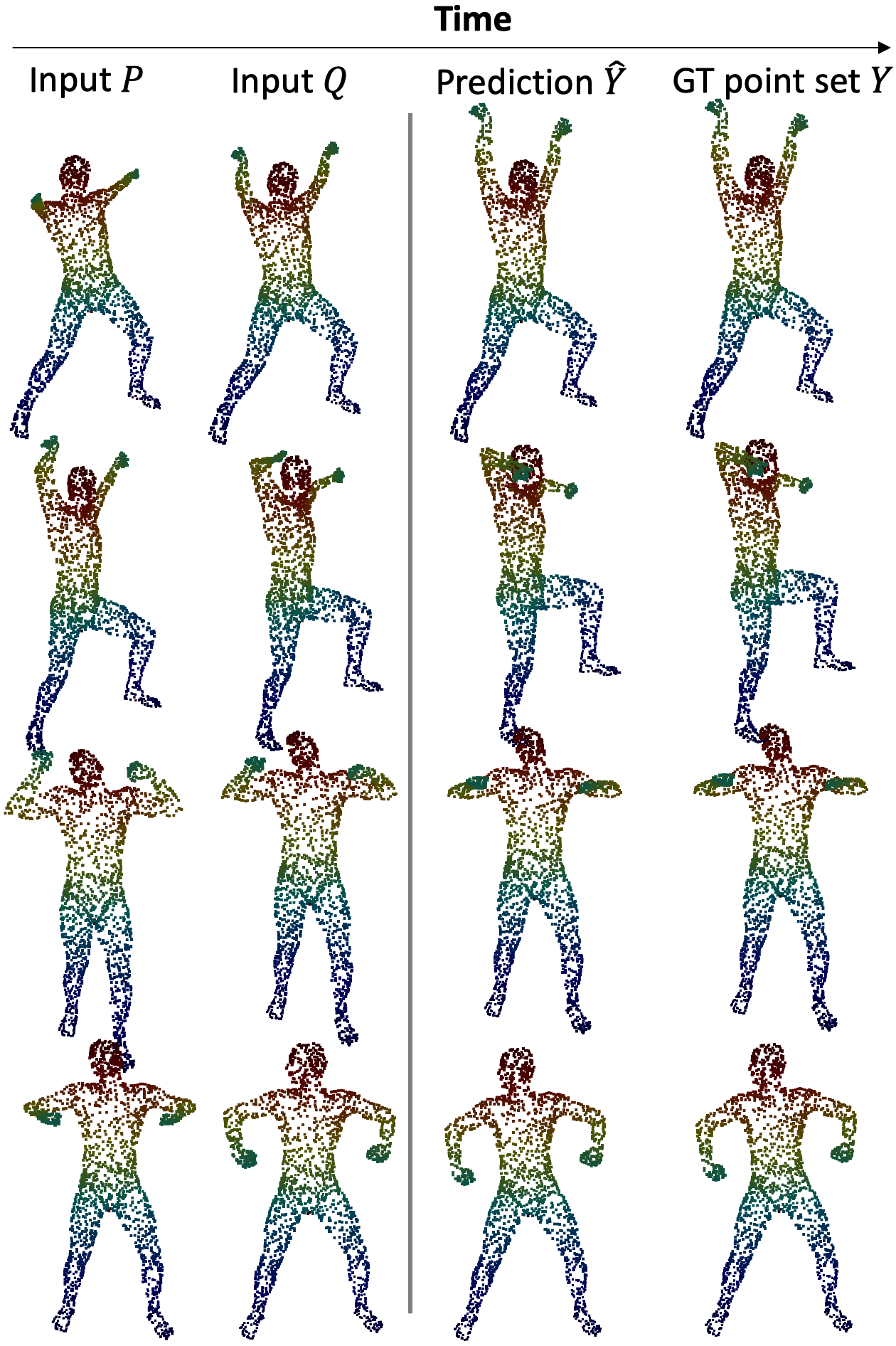}
\caption{Qualitative results on the temporal consistent SCAPE data set. Corresponding points are painted with same color.
\label{fig:scape}}
\end{figure}
\noindent\textbf{Experiment Settings: }
To verify the effectiveness of our proposed method, we first conduct the experiment on the temporally consistent data on SCAPE and D-Faust data set. We follow the training and testing procedure provided in the method part. We tested the performance of our model with different sizes of latent code $z$. Experimental results show that different sizes of z have a small effect on the final performance. We set the dimension of $z$ to 256 in the following sections.

\begin{figure}[!htp]
\centering
\includegraphics[width=0.8\linewidth]{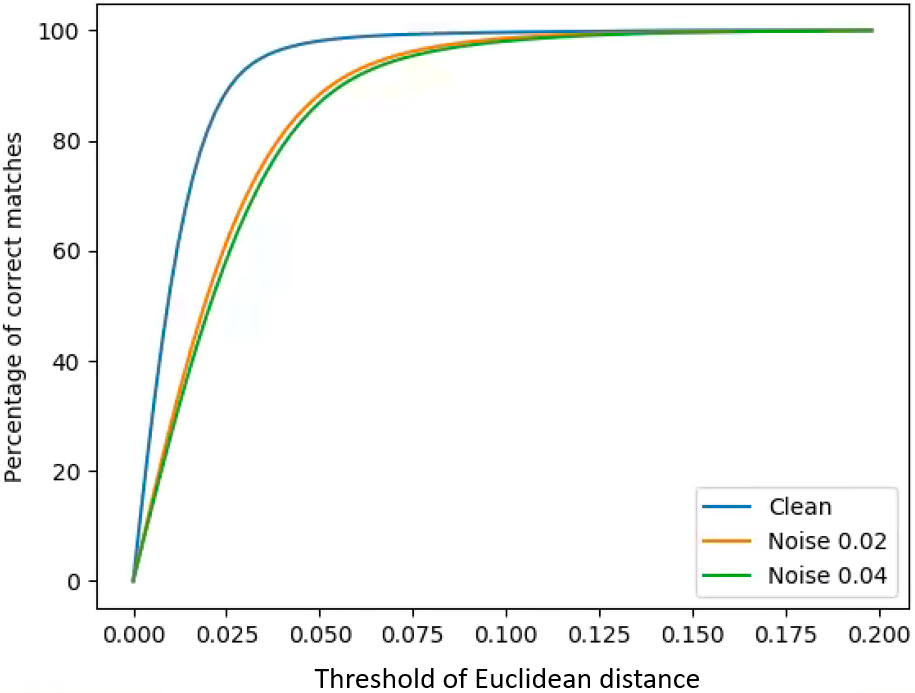}
\caption{Cumulative matching accuracy of our approach when taking different noise level of SCAPE data as inputs
\label{fig:noise}}
\end{figure}

\noindent\textbf{Result Analysis: } 
The visualized result on consistent inputs can be found in Fig.\ref{fig:scape}. The qualitative of our approach on both SCAPE and D-Faust dataset can be found in Table.\ref{tab:clean} and the blue line in the Fig.\ref{fig:noise}. As shown in this figure, our method can successfully predict the 3D motion from one frame to the next with satisfying correspondences. %To the best of our knowledge, we are the first work targeting the 3D motion prediction. 

\begin{table}[!htp]
\centering
\begin{tabular}{cccccccc}
\hline\hline
\multicolumn{8}{c}{3DMotion-Net}  
\\\hline
\multicolumn{4}{c|}{Data Set} &\multicolumn{2}{c|}{Chamfer Distance} &\multicolumn{2}{c}{Correspond. Distance}
\\\hline

\multicolumn{4}{l|}{SCAPE} &\multicolumn{2}{c|}{0.021} &\multicolumn{2}{c}{0.088} \\

\multicolumn{4}{l|}{D-FAUST} &\multicolumn{2}{c|}{0.029} &\multicolumn{2}{c}{0.091} \\

\hline\hline

\end{tabular}
\caption{
Quantitative results on D-Faust dataset. Chamfer distance and correspondence $\ell_2$-distance are reported. (Lower is better)}
\label{tab:clean}
\end{table}

\subsection{Motion Prediction on Inconsistent data set}
\noindent\textbf{Experiment Settings: }

\begin{figure}[!h]
\centering
\includegraphics[width=0.8\linewidth]{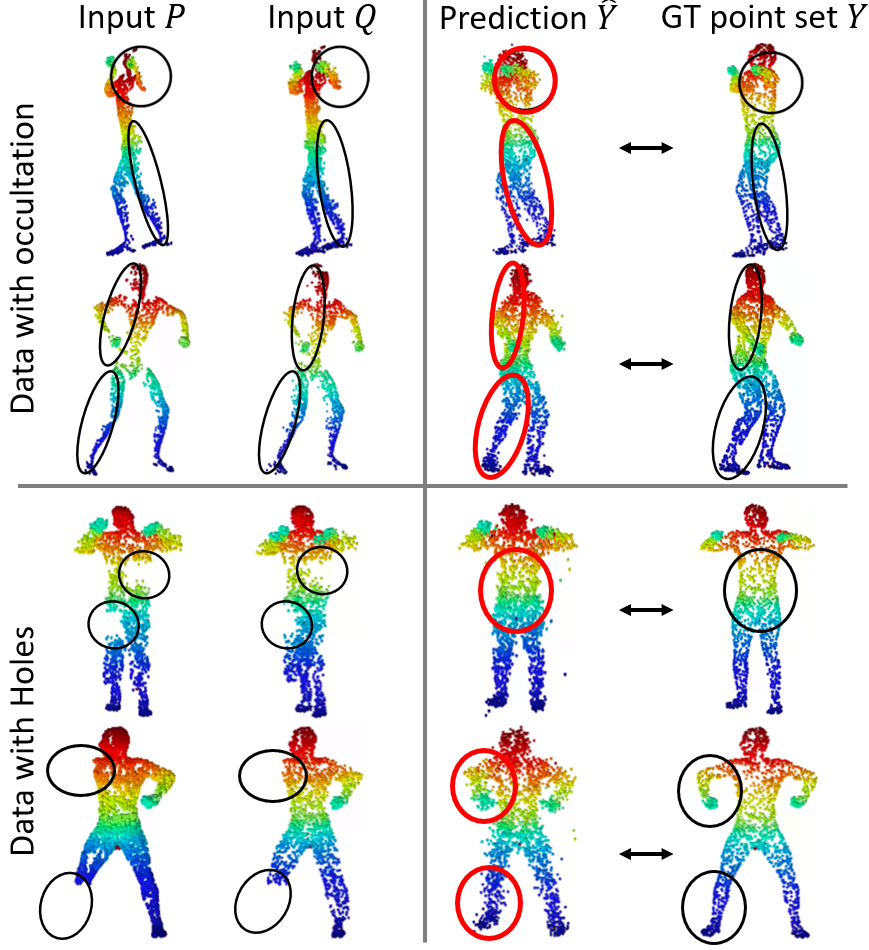}
\caption{Qualitative results on the inconsistent data with noises and occlusions. The high light part in this figure clearly shows that our proposed method can produce consistent prediction even when the inconsistent inputs are given.
\label{fig:scape_in}}
\end{figure}
3D scans are usually captured with data noise, partial observation, and holes. To show that our proposed can handle temporally inconsistent inputs, we first conduct experiments to test the noisy resistance ability by predicting future motion on the noisy motions at the different noise levels. Therefore, we add Gaussian noise to the SCAPE data set with noise levels of 0.02 and 0.04. Moreover, to further show the robustness of our model, we conduct an experiment with different input corruptions. We randomly cut off some parts of the object in the SCAPE data set to create the partial data set.

\begin{table}[!htp]
\centering
\begin{tabular}{cccccccc}
\hline\hline
\multicolumn{8}{c}{3DMotion-Net}  
\\\hline
\multicolumn{4}{c|}{Data Set} &\multicolumn{2}{c|}{Chamfer Distance} &\multicolumn{2}{c}{Correspond.}
\\\hline

\multicolumn{4}{l|}{SCAPE w/ Nosie Level 0.2} &\multicolumn{2}{c|}{0.029} &\multicolumn{2}{c}{0.091} \\

\multicolumn{4}{l|}{SCAPE w/ Nosie Level 0.4} &\multicolumn{2}{c|}{0.037} &\multicolumn{2}{c}{0.098} \\

\multicolumn{4}{l|}{SCAPE w/ Holes} &\multicolumn{2}{c|}{0.031} &\multicolumn{2}{c}{0.092} \\

\multicolumn{4}{l|}{SCAPE w/ Partial} &\multicolumn{2}{c|}{0.034} &\multicolumn{2}{c}{0.094} \\

\hline\hline

\end{tabular}
\caption{
Quantitative results on the inconsistent data. Chamfer distance and correspondence $\ell_2$-distance are reported. (Lower is better)}
\label{tab:in}
\end{table}

\noindent\textbf{Result Analysis: } 
During training process, our model is trained on the temporally consistent data and learns the prior information of how to decode latent codes into spatial consistent flow fields. This ensures our model can still search the optimal latent codes for motion prediction with inconsistent input data. From the Fig.\ref{fig:scape_in}, one can see that our method achieves a reasonable performance even with input noises and missing parts. This demonstrate that even when input data is inconsistent, our method can still use the designed loss to search the optimal latent code $z$ and to well predict the future motion. This can still be proved by the quantitative results in Table.\ref{tab:in}

\subsection{Cross-category Motion Prediction}
\noindent\textbf{Experiment Settings: }

\begin{figure}[!htp]
\centering
\includegraphics[width=0.9\linewidth]{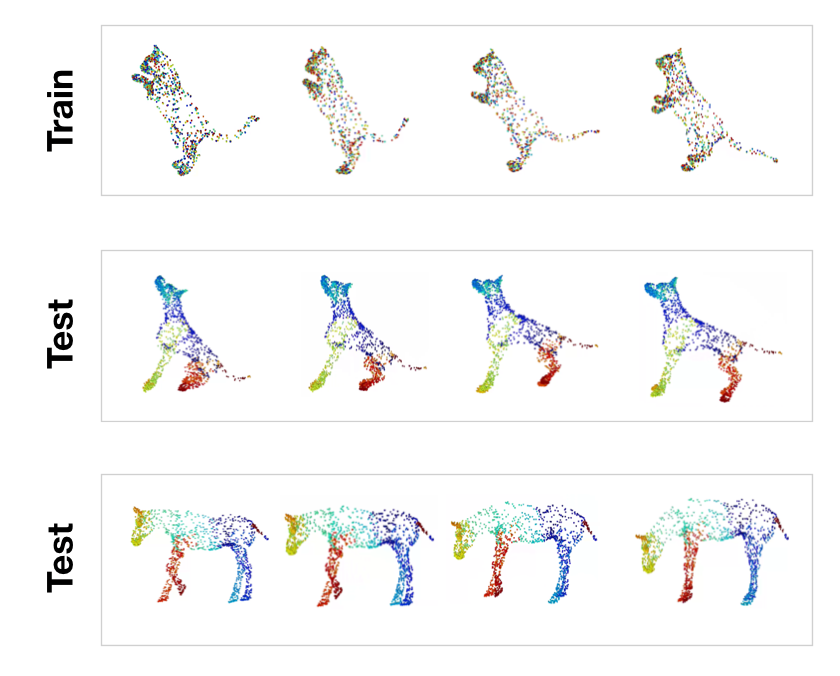}
\caption{Cross-category Motion Prediction. Our model is trained on cat category and evaluated on dog and horse categories. Corresponding
points are painted with same color.
\label{fig:tosca}}
\end{figure}

As described in the method section, our Temporal-aware shape descriptor (TaSD) is designed to learn temporal flow patterns between the current frame and the next frame regardless of the object categories. To verify the class transferring abilities of our proposed TaSD, we carry out further experiments on the TOSCA data set that we train our model on the cat category, and evaluate the motion prediction performance on other two categories, including dog and horse.

\noindent\textbf{Result Analysis: } 
As presented in the Figure.\ref{fig:tosca},all corresponding points are consistent across all the frames. As shown in Table.\ref{tab:tosca}, our model gets similar performance for horse and dog categories compared to the cat category. This demonstrates that our proposed TaSD is powerful to extract temporal flow patterns regardless the object category. %Moreover, as shown in Table\ref{tab:tosca}, the motion prediction performance on horse and dog categories are quite close. This is mainly because that cats, horses, and dogs are all “Four-Leg” animals that the degrees of the freedom for each joint point is the same.
Qualitative results in Figure.\ref{fig:tosca} also proves the same finding.

\begin{table}[!htp]
\centering
\begin{tabular}{cccccccc}
\hline\hline
\multicolumn{8}{c}{3DMotion-Net}  
\\\hline
\multicolumn{4}{c|}{Inference Category} &\multicolumn{2}{c|}{Chamfer Distance} &\multicolumn{2}{c}{Correspond.}
\\\hline

\multicolumn{4}{l|}{Cat (Training)} &\multicolumn{2}{c|}{0.020} &\multicolumn{2}{c}{0.087} \\

\multicolumn{4}{l|}{Dog (testing)} &\multicolumn{2}{c|}{0.023} &\multicolumn{2}{c}{0.090} \\

\multicolumn{4}{l|}{Horse (testing)} &\multicolumn{2}{c|}{0.023} &\multicolumn{2}{c}{0.089} \\

\hline\hline

\end{tabular}
\caption{
Quantitative results of the cross-category experiment. Our model is trained on TOSCA "CAT" category and the Chamfer distance and correspondence $\ell_2$-distance are reported. (Lower is better)}
\label{tab:tosca}
\end{table}

\section{Conclusion}
In this paper, we introduce a self-supervised method that can learn continuous flow function from two consecutive 3D scans. This function can predict the future object motion and naturally bring out the shape correspondences simultaneously by using the object prior and observed motion flow provided by the input two consecutive 3D scans. To eliminate the unsolved and challenging problem of defining a discrete point convolution on continuous space to encode spatial and temporal information, we introduce a temporal-aware motion descriptor to represent both temporal and geometric information. A shape morph decoder network is designed to learn a continuous motion flow function which morphs a 3D scan from the current frame to the next frame. A sequence of time-varying 3D scans is used to jointly train the decoder and temporally aware motion descriptor without the supervision of ground truth. Experiments on several benchmark data sets demonstrate a satisfying motion prediction performance of our model on both clean and inconsistent inputs.

\bibliographystyle{IEEEtran}
\bibliography{main}

\end{document}